# Grammaticality Representation in ChatGPT as Compared to Linguists and Laypeople

Zhuang Qiu[1], Xufeng Duan[1], and Zhenguang G. Cai[*,1,2]

[1] Department of Linguistics and Modern Languages, The Chinese University of Hong Kong
[2] Brain and Mind Institute, The Chinese University of Hong Kong

**Abstract**
Large language models (LLMs) have demonstrated exceptional performance across various linguistic tasks. However, it remains uncertain whether LLMs have developed human-like fine-grained grammatical intuition. This preregistered study (https://osf.io/t5nes) presents the first large-scale investigation of ChatGPT's grammatical intuition, building upon a previous study that collected laypeople's grammatical judgments on 148 linguistic phenomena that linguists judged to be grammatical, ungrammatical, or marginally grammatical (Sprouse, Schütze, & Almeida, 2013). Our primary focus was to compare ChatGPT with both laypeople and linguists in the judgement of these linguistic constructions. In Experiment 1, ChatGPT assigned ratings to sentences based on a given reference sentence. Experiment 2 involved rating sentences on a 7-point scale, and Experiment 3 asked ChatGPT to choose the more grammatical sentence from a pair. Overall, our findings demonstrate convergence rates ranging from 73% to 95% between ChatGPT and linguists, with an overall point-estimate of 89%. Significant correlations were also found between ChatGPT and laypeople across all tasks, though the correlation strength varied by task. We attribute these results to the psychometric nature of the judgment tasks and the differences in language processing styles between humans and LLMs.

## 1 Introduction

The technological progression within artificial intelligence, especially when it comes to the realm of natural language processing, has ignited significant discussions about how closely large language models (LLMs), including chatbots like ChatGPT, emulate human linguistic cognition and utilization (Chomsky, Roberts, & Watumull, 2023; Piantadosi, 2023; Binz and Schulz; 2023). With each technological leap, distinguishing between human linguistic cognition and the capabilities of AI-driven language models becomes even more intricate (Wilcox et al., 2022; Van Schijndel & Linzen, 2018; Futrell et al., 2019). This leads scholars to query if these LLMs genuinely reflect human linguistic nuances or merely reproduce them on a cosmetic level (Cai et al., 2023). This research delves deeper into the congruencies and disparities between LLMs and humans, focusing primarily on their instinctive understanding of grammar. In three preregistered experiments, ChatGPT was asked to provide grammaticality judgement in different formats for over two thousand sentences with diverse structural configurations. We compared ChatGPT's judgements with judgements from laypeople and linguists to map out any parallels or deviations.

The ascent of LLMs has been nothing short of remarkable, displaying adeptness in a plethora of linguistic challenges, including discerning ambiguities (Ortega-Martín, 2023), responding to queries (Brown et al., 2020), and transcribing across languages (Jiao et al., 2023). Interestingly, while these models weren't inherently designed with a hierarchical syntactical structure specifically for human languages, they have shown the capability to discern complex filler-gap dependencies and develop incremental syntactic interpretations (Wilcox et al., 2022; Van Schijndel & Linzen, 2018; Futrell et al., 2019). But the overarching question lingers: Do LLMs genuinely mirror humans in terms of linguistic cognition? Chomsky, Roberts, and Watumull (2023) have been vocal about the inherent discrepancies between how LLMs and humans perceive and communicate. Yet, other scholars like Piantadosi (2023) hold a contrasting view, positioning LLMs as genuine reflections of human linguistic cognition.

---

[*] Correspondence should be addressed to Zhenguang G. Cai (Department of Linguistics and Modern Languages, The Chinese University of Hong Kong, Leung Kau Kui Building, Shatin, Hong Kong SAR; email: zhenguangcai@cuhk.edu.hk).

Empirical studies have emerged as a crucial tool to answer this debate. Pioneering work by Binz and Schulz (2023) subjected GPT-3 to a battery of psychological tests, originally crafted to understand facets of human thought processes, ranging from decision-making matrices to reasoning pathways. The outcomes were intriguing, with GPT-3 not just mirroring but at times outperforming human benchmarks in specific scenarios. On a similar trajectory, Kosinski (2023) assessed the capacity of LLMs to understand and respond to false-belief scenarios, often utilized to gauge human empathy and comprehension. Here, the responses from ChatGPT echoed the patterns seen in school-going children, though subsequent research from Brunet-Gouet and colleagues (2023) voiced concerns about the consistency of such responses. Delving into ChatGPT's language processing abilities, Cai et al. (2023) subjected ChatGPT to a myriad of psycholinguistic experiments and showed an impressive alignment between the models and humans in language use in a majority of the tests, ranging from sounds, to syntax, all the way to dialogue. However, it's noteworthy that ChatGPT can diverge from humans in language use, for example, in word length preference for conveying lesser information (e.g., Mahowald et al., 2013).

When examining LLM-human similarities, it's crucial to assess the extent to which ChatGPT's representations of linguistic knowledge align with those of humans. Contemporary linguistic theories often distinguish between the inherent mental systems that enable language comprehension and production, and the actual use of language—illustrated by distinctions like "Langue vs. Parole" from Saussure (1916) and "Competence vs Performance" by Chomsky (1965). Grammaticality judgement is a central method to assess linguistic representation competence. Chomsky (1986) highlighted that evidence for linguistic theorizing largely depends on "the judgements of native speakers". While there are other sources of evidence, like speech corpus or acquisition sequences (Devitt, 2006), formal linguists typically favor native speakers' grammaticality intuitions. The prevailing assumption is that our language knowledge comprises abstract rules and principles, forming intuitions about sentence well-formedness (Graves et al. 1973; Chomsky 1980; Fodor 1981). However, relying on grammaticality judgments to frame linguistic theories isn't without dispute. Hill (1961) noted that such judgments often disregard acoustic properties like intonations, potentially compromising informant reliability. The dual role of formal linguists, as both theory developers and data providers, might compromise objectivity (Lyons, 1968; Ferreira, 2005). However, advancements have been made, including better practices for eliciting judgement data (Schütze, 1996), improving the reliability and validity of grammaticality judgements. Our study doesn't evaluate these methods, but we embrace grammaticality judgement for studying LLM knowledge representation, deeming it a practical tool. This decision rests on several reasons. First of all, more objective or direct measure of linguistic competence is not available for a comparative study between LLMs and human participants. Furthermore, generative linguistics' explanatory and predictive power attests to the value of metalinguistic judgments (Riemer, 2009). Empirical studies also affirm the reliability of controlled grammaticality judgment tasks (Langsford et al, 2018).

Formal surveys assessing sentence grammaticality often take the form of acceptability judgment tasks. Rather than asking participants to determine if a sentence is "grammatical", researchers frequently inquire whether sentences under consideration are "acceptable" (Sprouse, Schütze, & Almeida, 2013), "sound good" (Davies & Kaplan, 1998; van der Lely, Jones & Marshall, 2011), or are "possible" (Mandell, 1999). Chomsky (1965) elucidated the conceptual differences between grammaticality and acceptability. Here, "grammaticality" pertains to linguistic competence, while "acceptability" addresses the actual use of language. Acceptability is influenced not only by grammaticality but also by factors such as "memory limitations, intonational patterns, and stylistic considerations" (p. 11). As an illustration, slang, though grammatically correct, might be inacceptable in formal settings. In such instances, the demarcation between grammaticality and acceptability, based on social norms, is evident. In other scenarios, distinguishing between the two becomes ambiguous. For instance, multiple center-embedding sentences like "The rat the cat the dog chased killed ate the malt" are generally perceived as complex or challenging to interpret. Yet, the debate persists whether they should be categorized as "ungrammatical and unacceptable" or "grammatical but unacceptable" (Chomsky 1965; Bever, 1968). A consensus among researchers is that the generative concept of grammaticality is a subconscious mental representation. Therefore, in their pursuit of formulating a mental grammar theory, they operationalize grammaticality as acceptability (Riemer, 2009). Echoing Schütze (1996), we view grammaticality judgment and acceptability judgment synonymously, both gauging informants' intuition of sentence "goodness" from a grammatical perspective, as opposed to alignment with socio-cultural norms.

Sprouse, Schütze, and Almeida (2013) surveyed 936 participants to obtain their judgments on the grammatical acceptability of various English sentences across three tasks. These sentences, exemplifying 148 pairwise syntactic phenomena, were sampled from the journal *Linguistic Inquiry*, with eight sentences representing each phenomenon. Though linguists had previously classified these sentences as grammatical, ungrammatical, or marginally grammatical, the aim of Sprouse and colleagues was to determine the degree of convergence between laypeople's formal judgments and those of linguistic experts. They recruited native English speakers online for three distinct judgment tasks. In the Magnitude Estimation Task (ME task), participants were given a reference sentence with a pre-assigned acceptability rating. They were then asked to rate target sentences using multiples of the reference rating. In the 7-point Likert Scale Judgment Task (LS task), participants rated the grammatical acceptability of target sentences on a 7-point scale from least to most acceptable. Finally, in the two-alternative forced-choice task (FC task), participants were shown a pair of sentences (one deemed more grammatical than the other by linguists) and were asked to select the more grammatically acceptable option.

The collected data was analyzed using four statistical tests for each pairwise phenomenon. In particular, Sprouse et al. (2013) evaluated if laypeople rated grammatical sentences more favorably than their ungrammatical counterparts, as predicted by linguists. After summarizing the results of all 148 phenomena, the team calculated the convergence rate between expert informal ratings and laypeople's formal ratings. They identified a 95% convergence rate, implying that both linguists and laypeople generally agreed on sentence grammaticality 95% of the time. As one of the pioneering large-scale surveys on the influence of research paradigms on grammaticality judgment, Sprouse et al. (2013) not only affirmed the legitimacy of expert ratings but also endorsed the three judgment tasks, all later corroborated as reliable grammatical knowledge measures by subsequent studies (e.g., Langsford et al., 2018).

The crux of our study revolves around the representation of grammatical well-formedness in LLMs like ChatGPT. Learners who utilize LLMs for writing assistance often assume that these models possess expert-level grammatical knowledge of the target languages (Wu et al., 2023). Similarly, researchers conducting experiments initially designed for humans (as seen in Cai et al., 2023; Binz & Schulz, 2023) expect LLMs to interpret written instructions similarly to native speakers. Yet, there is limited research into the grammatical intuitions of LLMs like ChatGPT, especially in comparison to the judgments of both linguists and laypeople across a broad range of grammatical phenomena. In this paper, we present a comprehensive exploration of ChatGPT's grammatical intuition. Using the acceptability judgment tasks from Sprouse et al. (2013), ChatGPT evaluated the grammaticality of 2,355 English sentences across three preregistered experiments (https://osf.io/t5nes). Its judgment patterns were juxtaposed against those of both laypeople and linguists. Our findings indicate a substantial agreement between ChatGPT and humans regarding grammatical intuition, though certain distinctions were also evident.

## 2 Experiment 1

In this experiment, we presented ChatGPT with a reference sentence that had a pre-assigned acceptability rating of 100. We then asked ChatGPT to assign a rating, in multiples of this reference rating, to target sentences. This approach is a replication of the ME task from Sprouse et al. (2013), but with two key modifications. First, rather than involving human participants, we sourced judgment data directly from ChatGPT. Second, our data collection adopted a "one trial per run" procedure, meaning each interaction session (or run) with ChatGPT encompassed only the instructions and a single experimental sentence (together with some filler sentences; see below). This procedure was chosen to mitigate any influence previous trials might have on ChatGPT's subsequent judgments. We merged the human data from the ME task (available at https://www.jonsprouse.com/) with ChatGPT's judgment data, subsequently examining both convergences and divergences using a variety of statistical tests.

### 2.1 Method

Experimental items were adopted from the stimuli of Sprouse et al. (2013). These consisted of 2,355 English sentences that represented 148 pairwise syntactic phenomena, sampled from the Journal of *Linguistic Inquiry*[†]. Each phenomenon pair featured grammatically correct sentences and their less grammatical counterparts, which could either be outright ungrammatical or marginally so. Table 1 provides examples of these experimental items.

---

[†] Sprouse et al. (2013) initially sampled 150 pairwise phenomena. However, upon closer examination, they identified two duplicated pairs, resulting in 148 unique pairwise phenomena. While they incorporated 16 sentences for each pairwise phenomenon, some sentences were duplicated. Consequently, there were a total of 2,355 unique sentences in Sprouse et al. (2013).

| Pairwise Syntactic Phenomenon 1 | |
|---|---|
| **Grammatical Sentences** | **Ungrammatical Sentences** |
| It seems to him that Kim solved the problem. | He seems to that Kim solved the problem. |
| It appears to them that Chris is the right person for the job. | They appear to that Chris is the right person for the job. |
| It seems to her that Garrett should be punished for lying. | She seems to that Garrett should be punished for lying. |
| It appears to me that Dana is an unsafe driver. | I appear to that Data is an unsafe driver. |
| It seems to me that Robert can't be trusted. | I seem to that Robert can't be trusted. |
| It appears to her that Kyle cheated on his homework. | She appears to that Kyle cheated on his homework. |
| It seems to them that Sandra hates cooking. | They seem to that Sandra hates cooking. |
| It appears to him that Erin enjoys swimming. | He appears to that Erin enjoys swimming. |
| **Pairwise Syntactic Phenomenon 2** | |
| **Grammatical Sentences** | **Marginally Grammatical Sentences** |
| Ginny remembered to bring the beer. | Ginny remembered to have bought the beer. |
| Thomas tried to stop the thief. | Thomas tried to have stopped the thief. |
| Susan attempted to perform a backflip. | Susan attempted to have performed a backflip. |
| Bobby planned to attend college. | Bobby planned to have attended college. |
| Sarah hoped to go to the party. | Sarah hoped to have gone to the party. |
| Scott intended to run for class president. | Scott intended to have run for class president. |
| Vanessa refused to take out the garbage. | Vanessa refused to have taken out the garbage. |
| Michael managed to drive his car. | Michael managed to have driven his car. |

Table 1: Examples of experimental items from Sprouse et al. (2013) that were used in all three experiments reported in this paper.

We prompted ChatGPT to rate the grammatical acceptability of these items relative to the acceptability of a benchmark sentence. Following Sprouse, Wagers, and Phillips (2013), we used the sentence "Who said my brother was kept tabs on by the FBI?" as the reference sentence, assigning it an acceptability rating of 100. Adhering to our data collection procedure pre-registered with the Open Science Framework (https://osf.io/t5nes), we procured responses from the ChatGPT version dated Feb 13. In each run or session, a Python script mimicked human interaction with ChatGPT, prompting it to function as a linguist, evaluating the grammatical acceptability of sentences against the reference. Before rating experimental items, ChatGPT was exposed to six practice sentences, spanning various degrees of grammaticality. ChatGPT's responses were limited to numerical rating scores, without any supplementary comments or explanations. These responses were then logged for analysis.

> Dear ChatGPT, I would like you to serve as a linguist and assess the grammatical acceptability of a sentence in relation to the following reference sentence: "Who said my brother was kept tabs on by the FBI?" Let's assume the grammatical acceptability score for this reference sentence is 100. Your task is to assign a score to indicate how grammatically acceptable a new sentence is compared to the reference sentence. For example, if the new sentence sounds more grammatically acceptable than "Who said my brother was kept tabs on by the FBI?" then its rating score should be higher than 100 and vice versa. Please always refer to the reference sentence and its assigned score when evaluating the grammatical acceptability of a new sentence. Given that the reference sentence "Who said my brother was kept tabs on by the FBI?" has a score of 100, what grammatical acceptability score would you assign to each of the following sentences?
>
> 1. What kind of movies do you like the best?
> 2. Who do you like the best kind of fruits?
> 3. What city do you like best in Europe?
> 4. The cat is a domestic species of small carnivorous mammal.
> 5. The dog small the is cat species of big.
> 6. Cat is domestic species of small carnivorous mammal.
> 7. It seems to him that Kim solved the problem.
>
> Please ONLY give me seven scores in your response, without other words.

Figure 1: Example of a run of an experimental trial. In this trial, sentences 1 to 6 are practice sentences, while 7 is the experimental item.

Our data collection approach emphasized the "one trial per run" paradigm. In this mode, each ChatGPT interaction contained only a singular experimental trial. Contrary to the original procedure where each participant was given a 50-item survey, this method minimized potential biases stemming from preceding trials on ChatGPT's immediate judgment. This also circumvented an issue observed in prior projects, where ChatGPT would occasionally lose track of the instructions midway. Additionally, shorter sessions, characteristic of the "one trial per run" design, were less vulnerable to potential server or connectivity problems. In total, the experiment comprised 2,368 items: 2,355 unique sentences with 13 repetitions, in line with Sprouse et al. (2013). An example of one run of an experimental trial is in Figure 1. We conducted 50 experiment runs for each item.

We conducted two sets of statistical analyses to address two research questions. The first pertains to the degree to which ChatGPT demonstrates grammatical intuition comparable to that of human participants who are not necessarily linguists. To address this question, we integrated data sourced from ChatGPT with the human data. Following Sprouse et al. (2013), ratings were standardized by participants using z-score transformation. By-item mean ratings were calculated for each experimental item from both ChatGPT and human responses. A correlation analysis was then conducted based on these by-item means, with the coefficient indicating the degree of agreement between ChatGPT's grammatical intuition and that of the human participants. According to Cohen (1988, 1992), a correlation coefficient of 0.5 or higher is considered indicative of a strong correlation.

However, it is important to note that a strong correlation doesn't imply perfect equivalence. To determine if ChatGPT's grammatical knowledge differs from that of humans, we devised a Bayesian linear mixed-effects model using R package brms (Bürkner, 2017). In this model, the acceptability rating score is a function of grammaticality (grammatical vs.

ungrammatical), participant type (human vs. ChatGPT), and their interactions. Predictors were dummy-coded, with the baseline being ChatGPT's judgement of grammatical sentences. The model incorporated random effects structures for by-item intercepts and slopes as illustrated below:

score ~ grammaticality*participant type + (1 + grammaticality*participant type | item)

For this model, a main effect of grammaticality would be expected, given that grammatical sentences should generally have a higher acceptability rating than ungrammatical ones. If a significant main effect for participant type or an interaction effect emerges, this would suggest that ChatGPT and human participants have different grammatical knowledge. Conversely, the absence of such effects would imply comparable grammatical competence between ChatGPT and humans.

Our second objective was to determine the extent of ChatGPT's grammatical knowledge aligns with that of expert linguists. To this end, we computed the convergence rate between ChatGPT's judgements and experts' assessments, employing the same techniques used in Sprouse et al. (2013). For each pairwise phenomenon, five distinct analyses were performed on ChatGPT's rating data to ascertain if grammatical sentences were rated higher than their ungrammatical counterparts, as judged to be by linguists. The outcomes of these analyses for all 148 pairwise phenomena were summarized. The percentage of phenomena wherein grammatical sentences achieved higher ratings than ungrammatical ones was treated as the convergence rate between expert assessments and ChatGPT's evaluations. The five analyses for each pairwise phenomenon consisted of: 1) Descriptive directionality 2) One-tailed t-test 3) Two-tailed t-test 4) Mixed-effects model 5) Bayes factor analysis.

In the Descriptive Directionality analysis, the average rating scores of the grammatical sentences were juxtaposed with those of the ungrammatical ones. Should the average of the grammatical sentences surpass that of the ungrammatical ones in a particular pairwise phenomenon, it would be interpreted as a convergence between ChatGPT and linguists in their grammatical judgements for that phenomenon. Both the one-tailed and two-tailed t-tests examined the statistical significance of the difference in means between grammatical and ungrammatical sentences. A conclusion of convergence in acceptability judgements between ChatGPT and linguists for a given phenomenon would be made only if the average rating for grammatical sentences significantly exceeded that for ungrammatical ones. The mixed-effects models were constructed utilizing the R package lme4 (Bates et al., 2020), modeling the rating score as a function of grammaticality with items treated as random effects. The Bayes Factor analysis utilized a Bayesian version of the t-test (Rouder et al., 2009) facilitated by the R package BayesFactor (Morey et al., 2022). Scripts for these analyses are accessible via the Open Science Framework (https://osf.io/crftu/).

## 2.2 Results
We observed a robust correlation between the by-item rating scores of ChatGPT and humans ($r = 0.69$, $p < 0.001$). As illustrated in Figure 2, sentences deemed more grammatical by humans similarly received higher acceptability ratings from ChatGPT, and vice versa.

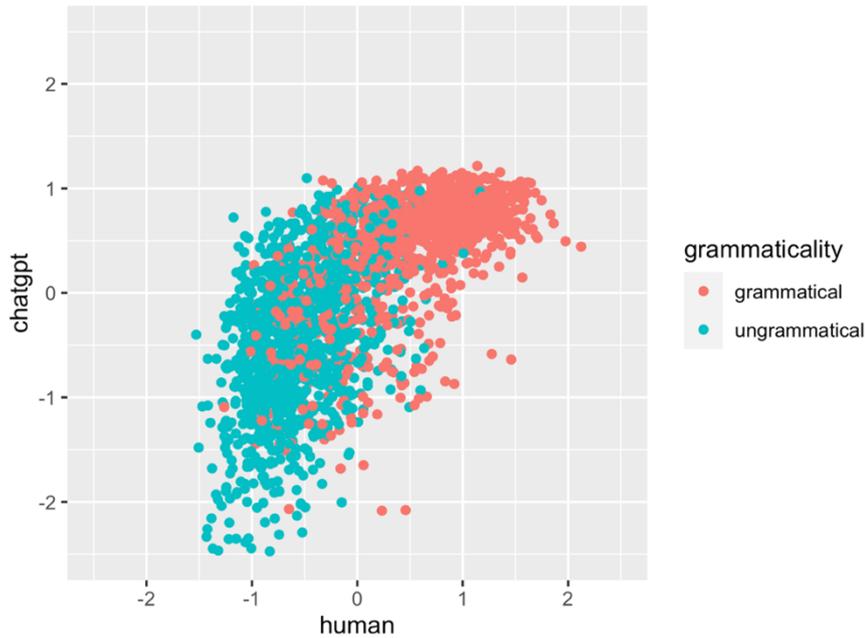

Figure 2: Correlation of acceptability ratings between human participants and ChatGPT in Exp1, with each point representing the mean rating score of a sentence.

To discern whether ChatGPT's ratings could be differentiated from those of human participants, we constructed a Bayesian linear mixed-effects model. In this model, the acceptability ratings were predicated on grammaticality, the participant type, and their interaction. Our findings showed a pronounced main effect of grammaticality: both ChatGPT and human participants rated ungrammatical sentences lower than their grammatical counterparts (see Figure 3 and Table 2). Interestingly, an interaction effect surfaced between participant type and grammaticality. For sentences that were grammatical, human participants awarded higher rating scores (0.07, CI = [0.04, 0.10]) compared to ChatGPT's ratings. Conversely, for ungrammatical sentences, humans attributed lower acceptability ratings (-0.15, CI = [-0.20, -0.10]) than ChatGPT.

|                      | Estimate | Est.Error | l-95% CI | u-95% CI |
|---------------------:|:--------:|:---------:|:--------:|:--------:|
| Intercept            | 0.46     | 0.02      | 0.42     | 0.49     |
| ungrammatical        | -0.91    | 0.03      | -0.97    | -0.85    |
| human                | 0.07     | 0.02      | 0.04     | 0.10     |
| ungrammatical: human | -0.15    | 0.03      | -0.20    | -0.10    |

Table 2: Summary of outputs from the Bayesian linear mixed-effects model in Exp1, using ChatGPT's judgment of grammatical sentences as the baseline for comparison. An estimate is considered statistically meaningful if the 95% credible interval does not include zero.

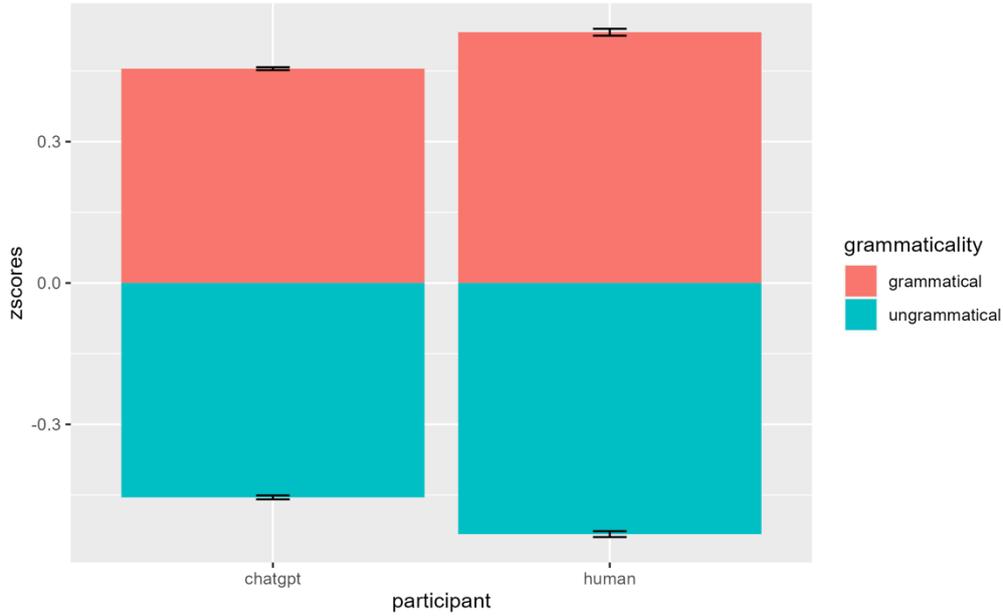

Figure 3: Comparison of average rating scores across participant types and grammaticality manipulation in Exp1. Error bars indicate standard errors.

Regarding the congruence between ChatGPT's ratings and linguists' judgments, 139 out of 148 pairwise phenomena showcased aligned directions. This indicates that for 139 of the 148 paired sets of sentences crafted by linguists, ChatGPT rated grammatical sentences as more acceptable than their ungrammatical counterparts when assessed solely by mean scores. The statistical significance of these mean differences was then evaluated using the methodologies outlined in Section 2.1, with the summary provided below.

|  | One-tailed | Two-tailed | LME | Bayes factor |
|---|---|---|---|---|
| Significant in the opposite direction | - | 7 | 0 | 7 |
| Marginal in the opposite direction | - | 0 | 2 | 0 |
| Non-significant in the opposite direction | - | 2 | 7 | 2 |
| Non-significant in the predicted direction | 10 | 5 | 25 | 5 |
| Marginal in the predicted direction | 4 | 0 | 6 | 3 |
| Significant in the predicted direction | 134 | 134 | 108 | 131 |

Table 3: Results of statistical tests assessing the convergence between ChatGPT and linguists in Exp1, using criteria from Sprouse et al. (2013). Note: Significant p-values are $p < .05$ and marginal p-values are $p \leq .1$. Significant Bayes factors are $BF > 3$, and marginal ones are $BF > 1$, in each direction.

The convergence rate estimates fluctuated based on the test applied. Both the classic null-hypothesis significance t-tests and the Bayesian t-test (Rouder et al., 2009) indicated higher convergence rates, ranging from 89% (131/148) to 91% (134/148). In contrast, linear mixed-effects models (LME) posited a lower convergence estimate of approximately 73% (108/148). Notably, while differences in mean ratings between grammatical and ungrammatical sentences were evident in ChatGPT's data, not all these differences were statistically significant. In certain instances, even when grammatical sentences held a higher average rating than ungrammatical ones, this differential lacked statistical significance, precluding it from being counted as a scenario where ChatGPT's judgment aligns with that of the linguists. These discrepancies across statistical tests are not unique to ChatGPT's dataset. As illustrated in Table 4, the convergence rate between laypeople and linguists also ranged from 86% (127/148) to 92% (136/148), contingent on the applied test.

|  | One-tailed | Two-tailed | LME | Bayes factor |
|---|---|---|---|---|
| Significant in the opposite direction | - | 2 | 2 | 2 |
| Marginal in the opposite direction | - | 0 | 0 | 0 |
| Non-significant in the opposite direction | - | 0 | 1 | 0 |
| Non-significant in the predicted direction | 11 | 10 | 16 | 14 |
| Marginal in the predicted direction | 1 | 4 | 2 | 2 |
| Significant in the predicted direction | 136 | 132 | 127 | 130 |

Table 4: Results of various statistical tests examining the convergence between laypeople and linguists, based on a re-analysis of human data from the ME task in Sprouse et al. (2013).

**2.3 Discussion**

In this experiment, we assessed the degree to which ChatGPT's grammatical intuition mirrors that of humans in the ME task. The outcomes responded directly to the research questions posed in Section 2.1. Firstly, a pronounced correlation emerged between ChatGPT's acceptability ratings and those of human participants who weren't necessarily linguistic experts. This correlation suggested that ChatGPT's capacity to discern grammatical acceptability resonates closely with judgments from human subjects, lending weight to the idea that ChatGPT, in spite of its AI origins, has linguistic intuitions akin to human language users.

Utilizing the Bayesian linear mixed-effects model, we gained a deeper understanding of the congruencies and disparities in the grammatical knowledge of ChatGPT and human participants. The aim was to ascertain whether distinctions could be drawn from their judgment patterns. Both cohorts consistently ranked grammatical sentences as more acceptable than their ungrammatical counterparts, thereby acknowledging a main effect of grammaticality. An interaction effect between participant type and grammaticality revealed subtle discrepancies in their acceptability ratings: humans tended to rate grammatical sentences higher than ChatGPT, while ChatGPT gave ungrammatical sentences higher ratings than humans did. These outcomes suggested that compared to humans, ChatGPT was more conservative in its ratings in the ME task.

The convergence analysis juxtaposing ChatGPT and linguistic experts shed light on the model's resonance with established linguistic judgments. Contingent upon the statistical methods applied, the estimated convergence rate for the ME task fluctuated between 73% and 91%. This implies that ChatGPT's grammatical determinations align substantially with expert linguistic judgments. Additionally, the ChatGPT-linguist convergence rate exhibited a broader range compared to the laypeople-linguist estimates (86% to 92%). However, given the unknown distribution of the convergence rate, its statistical significance remains ambiguous. This topic will be elaborated upon in the General Discussion section, where estimates from three experiments of this study are considered all together.

In our first experiment, ChatGPT's grammatical intuition was evaluated using the ME task. As per Langsford et al. (2018), ME scores can be influenced by individual response style variances, resulting in the between-participant reliability of the ME task being notably less than its within-participant reliability. This discrepancy isn't observed in other grammaticality measures like the Likert scale and force-choice tasks, where potential variations in response styles are minimized. To enhance our understanding, Experiment 2 and Experiment 3 implemented the Likert scale task and force-choice task to juxtapose ChatGPT's grammatical intuitions with those of both linguistic experts and laypeople.

## 3. Experiment 2

In this experiment, we further explored ChatGPT's grammatical intuition by comparing it with laypeople and linguists using the Likert scale task. This task offers greater within-participant and between-participant reliability than the ME task. The selection of sentence stimuli and the data-analysis procedures remained consistent with those of the first experiment.

### 3.1 Method

Experimental items were sourced from the same inventory of stimuli utilized in Experiment 1. However, instead of prompting ChatGPT to rate the grammatical acceptability of the experimental items relative to a reference sentence, we directly asked ChatGPT to denote the grammatical acceptability of each sentence on a 7-point Likert scale (1 = "least acceptable" and 7 = "most acceptable"). We employed the "one trial per run" data collection method, as in Experiment 1. Comprehensive details of the experimental procedure can be accessed in the preregistration report available at https://osf.io/t5nes. In total, 2,368 experimental items were tested, with each item undergoing 50 experimental runs.

### 3.2 Results

Figure 4 illustrates a strong correlation between by-item rating scores of ChatGPT and humans ($r = 0.72$, $p < 0.001$). Figure 5 shows that when comparing average rating scores across participant types and grammaticality manipulations, grammatical sentences consistently received notably higher acceptability ratings than ungrammatical ones. Notably, there was minimal difference between human ratings and those provided by ChatGPT.

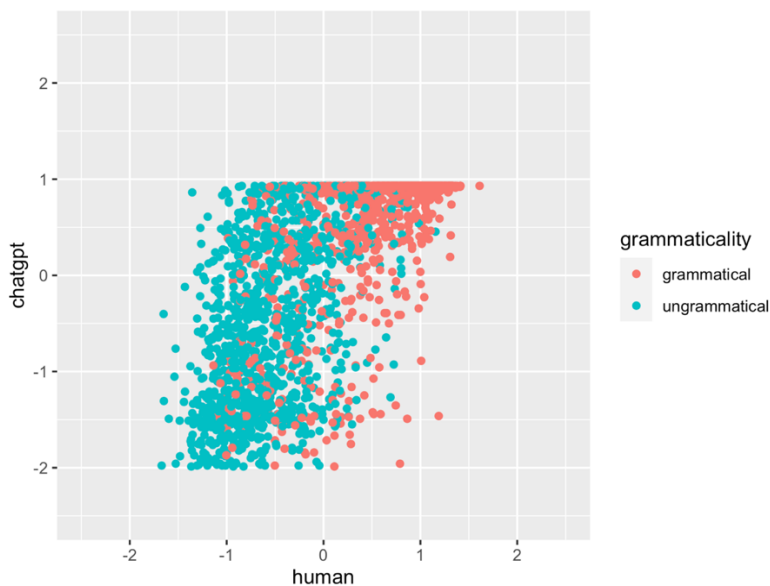

Figure 4: Correlation of acceptability ratings between human participants and ChatGPT in Exp2, with each point representing the mean rating score of a sentence.

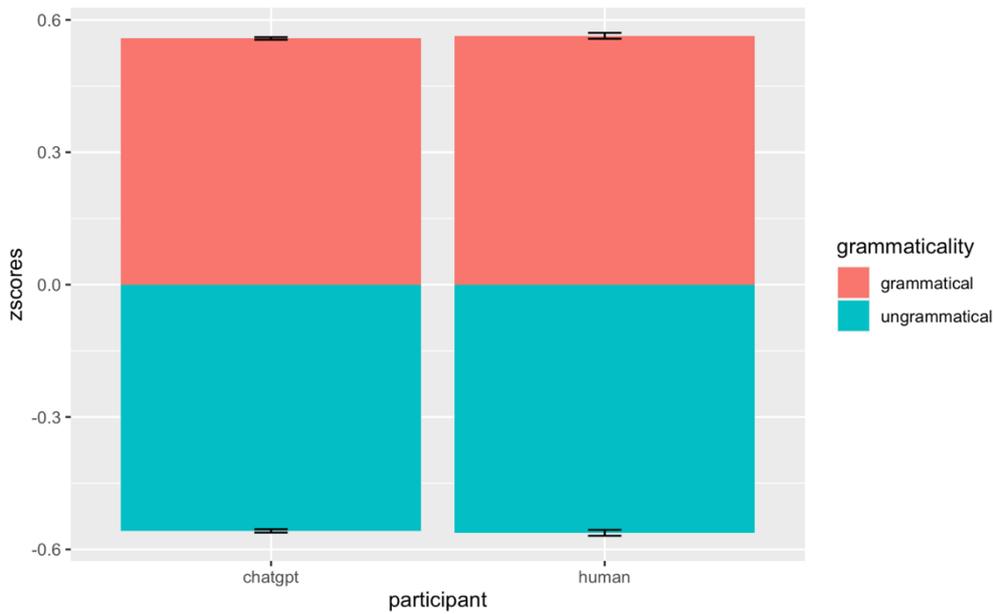

Figure 5: Comparison of average rating scores across participant types and grammaticality manipulation in Exp2. Error bars represent standard errors.

This observed trend was further corroborated by a Bayesian linear mixed-effects model. In this model, acceptability ratings were framed as a function of grammaticality, participant type, and their potential interactions. The main effect of grammaticality was apparent (-1.12, CI = [-1.18, -1.05]), but neither the effect of participant types (0.00, CI = [-0.03, 0.04]) nor their interactions (-0.01, CI = [-0.06, 0.05]) were significant.

|                      | Estimate | Est.Error | l-95% CI | u-95% CI |
|---------------------:|---------:|----------:|---------:|---------:|
| Intercept            | 0.56     | 0.02      | 0.52     | 0.60     |
| ungrammatical        | -1.12    | 0.03      | -1.18    | -1.05    |
| human                | 0.00     | 0.02      | -0.03    | 0.04     |
| ungrammatical: human | -0.01    | 0.03      | -0.06    | 0.05     |

Table 5: Summary of outputs from the Bayesian linear mixed-effects model in Exp2. An estimate is considered statistically meaningful if the 95% credible interval does not include zero.

Examining the convergence rate between ChatGPT's ratings and the judgments of linguists revealed a remarkable alignment. Specifically, for 141 out of the 148 pairwise phenomena, ChatGPT's ratings paralleled the direction of linguists' judgments. This indicates that in 141 out of 148 instances where linguists constructed pairs of grammatical and ungrammatical sentences, ChatGPT's acceptability ratings for grammatical sentences surpassed those of ungrammatical sentences—this observation holds when considering mean ratings alone. To further evaluate the significance of these mean differences, we employed methodologies delineated in Section 2.1, and the results were summarized below.

|                                           | One-tailed | Two-tailed | LME | Bayes factor |
|------------------------------------------:|:----------:|:----------:|:---:|:------------:|
| Significant in the opposite direction     | -          | 7          | 2   | 7            |
| Marginal in the opposite direction        | -          | 0          | 1   | 0            |
| Non-significant in the opposite direction | -          | 0          | 4   | 0            |
| Non-significant in the predicted direction| 8          | 1          | 23  | 2            |
| Marginal in the predicted direction       | 0          | 1          | 7   | 0            |
| Significant in the predicted direction    | 140        | 139        | 111 | 139          |

Table 6: Results of various statistical tests assessing the convergence between ChatGPT and linguists in Exp2, using criteria from Sprouse et al. (2013). Note: Significant p-values are p < .05 and marginal p-values are p ≤ .1. Significant Bayes factors are BF > 3, and marginal ones are BF > 1, in each direction.

Echoing findings from Experiment 1, the choice of significance test affected the estimated convergence rates. Classic null-hypothesis significance t-tests and Bayesian t-tests both inferred a higher convergence rate, ranging from 94% (139/148) to 95% (140/148). In contrast, linear mixed-effects models suggested a slightly more conservative convergence rate estimate of approximately 75% (111/148). For comparison's sake, when estimating the convergence rate between laypeople and linguists, figures varied between 87% (129/148) and 93% (137/148), contingent upon the chosen statistical test.

|                                           | One-tailed | Two-tailed | LME | Bayes factor |
|------------------------------------------:|:----------:|:----------:|:---:|:------------:|
| Significant in the opposite direction     | -          | 2          | 0   | 2            |
| Marginal in the opposite direction        | -          | 0          | 1   | 0            |
| Non-significant in the opposite direction | -          | 3          | 4   | 3            |
| Non-significant in the predicted direction| 10         | 6          | 13  | 10           |
| Marginal in the predicted direction       | 1          | 4          | 1   | 3            |
| Significant in the predicted direction    | 137        | 133        | 129 | 130          |

Table 7: Results of various statistical tests examining the convergence between laypeople and linguists, based on a re-analysis of human data from the LS task in Sprouse et al. (2013)

### 3.3 Discussion

In this experiment, we delved deeper into ChatGPT's ability to evaluate the grammatical acceptability of English sentences using a 7-point Likert scale task. The results once again revealed a robust correlation between acceptability ratings from ChatGPT and human participants, highlighting a strong agreement in their evaluations of grammatical sentences. Intriguingly, our statistical analysis did not find any significant differences based on participant type (human vs. ChatGPT) or interactions between participant type and grammaticality. This implies that, at least in the context of this Likert scale task, ChatGPT's grammatical competence aligns closely with that of human participants. Such findings emphasize the model's remarkable capacity to discern subtle differences in sentence acceptability, paralleling human linguistic judgments.

Beyond comparing ChatGPT with human evaluations, we examined its alignment with judgments from linguistic experts. Across 148 pairwise linguistic phenomena, the convergence rate underscored a considerable alignment between ChatGPT's ratings and those of linguists, ranging from 75% to 95%. This suggests that, in 75% to 95% of instances,

ChatGPT consistently ranked grammatical sentences as more acceptable than their ungrammatical counterparts, echoing the intuitions of linguistic professionals. While the range in convergence rate estimates across different statistical methodologies is noteworthy, it likely stems from the intrinsic characteristics of the statistical methods. Regardless, the consistently high convergence rate across different tests underscores ChatGPT's grammatical intuition in alignment with expert perspectives.

In our earlier experiments, we evaluated ChatGPT's grammatical knowledge using two distinct rating tasks: one asked ChatGPT to evaluate the acceptability of target sentences against a reference sentence, and the other used a 7-point Likert scale. While rating tasks offer participants the latitude to distinguish varying degrees of grammaticality, they also introduce questions concerning the consistency of participants in correlating numerical scores with perceived acceptability. For instance, does a sentence scored at five genuinely seem less acceptable than a sentence rated six later in the same experiment? Given this, it's paramount to supplement the rating tasks with a forced-choice task (FC task). This format affords participants less interpretative flexibility and thereby ensures heightened within-participant reliability.

## 4 Experiment 3

In this experiment, we replicated the FC task in Sprouse et al. (2013) with ChatGPT serving as the participant. Instead of presenting a sentence for an acceptability rating, we prompted ChatGPT to select the more grammatically acceptable sentence from a pair. We then compared ChatGPT's choices to those made by humans to assess similarities in their representation of grammatical knowledge.

### 4.1 Method

The experimental items were sourced from the same inventory of pairwise phenomena as used in Experiments 1 and 2. However, in this experiment, rather than presenting individual sentences separately from their paired counterparts, we displayed one grammatically correct sentence and its less grammatical counterpart in vertically arranged pairs. ChatGPT was prompted to indicate which of the two sentences in each pair was more acceptable. To counterbalance the order of grammaticality, for each phenomenon (such as Pairwise Syntactic Phenomenon 1 in Table 1), the grammatical sentence appeared above its less grammatical counterpart in four vertical pairs. Conversely, in the other four pairs, the less grammatical sentence was positioned above its grammatical mate. In total, Experiment 3 comprised 1184 trials, with an example of a trial presented as follows.

> Dear ChatGPT, I would like you to serve as a linguist and assess the grammatical acceptability of some sentences. I want you to read pairs of sentences, and for each pair, you need to tell me which of the two sentences is grammatically more acceptable. For each pair of the sentences, the first sentence you read is sentence 1 and the second sentence you read is sentence 2.
>
> sentence 1: It seems to him that Kim solved the problem.
> sentence 2: He seems to that Kim solved the problem.
>
> Which of the two sentences is grammatically more acceptable, sentence 1 or sentence 2? Please just say "sentence 1" or "sentence 2" without other words.

Figure 6: Example of a trial in Experiment 3

In a manner consistent with Experiment 1, we adhered to the preregistered data collection procedure (https://osf.io/t5nes) to obtain responses from ChatGPT (Feb 13 version). For each run (or experimental session), a Python script was used to simulate a human user interacting with ChatGPT. We instructed ChatGPT to assume the role of a linguist and choose the more grammatically acceptable sentence from each pair. Each experimental session presented only one experimental item, and we executed 50 runs for each item.

Although we instructed ChatGPT to reply with either "sentence 1" or "sentence 2" exclusively, ChatGPT occasionally deviated from this guideline. In some instances, ChatGPT either judged both sentences to be equally grammatical or provided extraneous information. In line with the preregistered data exclusion criteria, we excluded all responses not adhering to the instruction. The remaining responses were coded as "1" if ChatGPT correctly identified the more acceptable sentence in the pair, and "0" otherwise.

The first suite of statistical analyses aimed to discern the degree to which ChatGPT's grammatical intuition mirrored that of laypeople. We merged the data obtained from ChatGPT with human data from the FC task in Sprouse et al. (2013). Responses in the human dataset were re-coded as "1" or "0", akin to the ChatGPT data. For the integrated dataset, we modelled the logit (log-odds) of selecting the correct answer as a function of participant type (human vs. ChatGPT) using a Bayesian linear mixed-effects model. With ChatGPT as the baseline, the predictor was dummy coded, and the random effects structures incorporated by-item intercepts and slopes, illustrated in the subsequent formula:

accuracy ~ participant type + (1 + participant type | item)

Furthermore, we estimated the probability of choosing the grammatical sentence for each experimental item for both ChatGPT and human participants, subsequently conducting a correlational analysis between the two sets.

Echoing the methodologies of Experiment 1 and 2, we examined the extent of alignment between ChatGPT's grammatical knowledge and that of the linguists. For each pairwise phenomenon in the ChatGPT dataset, we employed five distinct analyses, summarizing instances where grammatical sentences held preference over their less grammatical counterparts. The quintet of analyses for each pairwise phenomenon included: 1) descriptive directionality, 2) one-tailed sign test, 3) two-tailed sign test, 4) mixed-effects model, and 5) Bayes factor analysis. These assessments mirrored those of Experiments 1 and 2 but were tailored for binary data. For instance, while Experiments 1 and 2 analysed descriptive directionality by juxtaposing the mean rating scores of grammatical and ungrammatical sentences, this experiment achieved the same by contrasting the selection proportion of grammatical versus less grammatical sentences. For this experiment, mixed-effects models were formulated as logistic regressions using the R package lme4. We calculated the logit of selecting grammatical sentences via an intercept-only model incorporating by-item random intercepts. Lastly, Bayes factor analyses, the Bayesian counterparts of the sign tests, were executed using the R package BayesFactor (Morey et al., 2022).

### 4.2 Results
Out of the 59,200 observations recorded for ChatGPT responses, 709 (or 1%) were excluded due to non-conformance to instructions. This left 58,491 (or 99%) of the data points for subsequent analysis. As illustrated in Figure 7, ChatGPT exhibited high accuracy in the FC task, comparable to human participants. The proportion of correct responses for human participants was 89%, slightly surpassing ChatGPT's 88%. However, when considering the random effects of trials, the mixed-effects model indicated that human participants (beta = -18.12, CI = [-21.89, -15.03], as detailed in Table 8) were somewhat less accurate than ChatGPT.

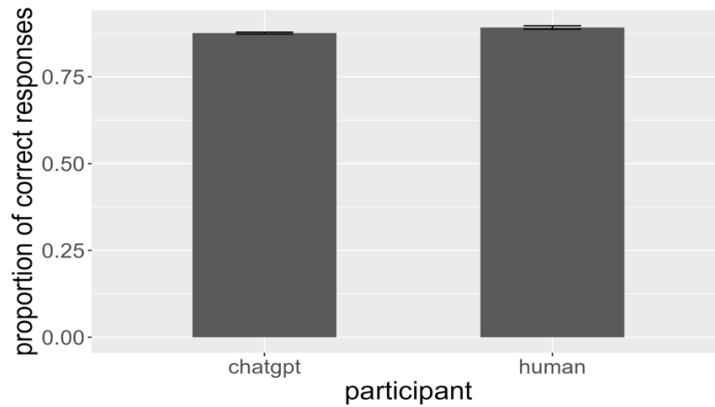

Figure 7: Comparison of the proportion of correct responses between ChatGPT and human participants across all trials in Experiment 3. Error bars represent confidence intervals.

|  | Estimate | Est.Error | l-95% CI | u-95% CI |
|---|---|---|---|---|
| Intercept | 21.24 | 1.74 | 18.13 | 25.03 |
| human | -18.12 | 1.73 | -21.89 | -15.03 |

Table 8: Summary of the outputs from the Bayesian linear mixed-effects model in Exp3. An estimate is considered statistically meaningful if the 95% credible interval does not include zero.

We performed a correlation analysis between human participants and ChatGPT concerning the probability of choosing grammatical sentences in each trial. The results revealed a modest correlation between the two groups (r = 0.39, p < 0.001, see Figure 8).

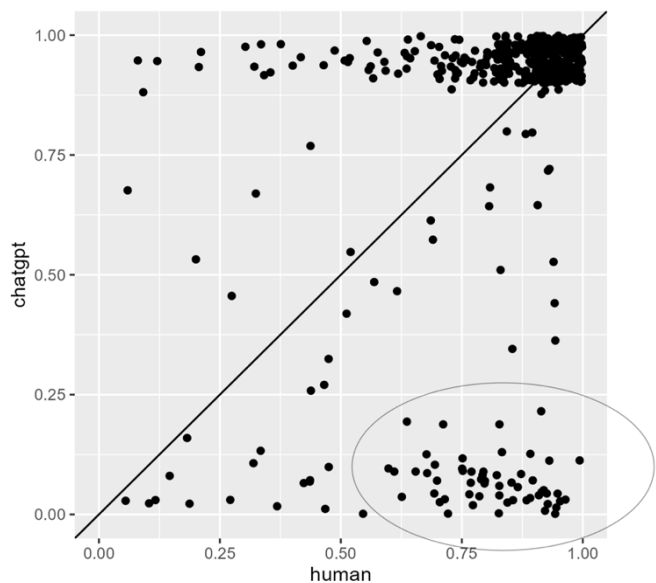

Figure 8: Probability of selecting grammatical sentences in the FC task for each experimental item, comparing human participants and ChatGPT. A diagonal reference line is added for ease of comparison. Data points above the line indicate trials where ChatGPT outperformed humans, and those below show where humans had higher accuracy. Outliers are circled.

In evaluating the alignment between ChatGPT's choices and linguists' judgements, we observed that for 133 of the 148 pairwise phenomena, ChatGPT's judgements mirrored those from the linguists. This indicates that in 133 of the 148 sets of grammatical and ungrammatical sentence pairs designed by linguists, ChatGPT accurately deemed the grammatical sentences as more acceptable than their less grammatical counterparts. We further probed the significance of the preference for grammatical sentences in each pairwise phenomenon using the methodologies outlined in Section 4.1, and the outcomes are summarized below.

|  | One-tailed | Two-tailed | LME | Bayes factor |
| --- | --- | --- | --- | --- |
| Significant in the opposite direction | - | 10 | 6 | 7 |
| Marginal in the opposite direction | - | 1 | 1 | 3 |
| Non-significant in the opposite direction | - | 4 | 8 | 5 |
| Non-significant in the predicted direction | 18 | 3 | 2 | 3 |
| Marginal in the predicted direction | 0 | 0 | 0 | 0 |
| Significant in the predicted direction | 130 | 130 | 131 | 130 |

Table 9: Results of various statistical tests assessing the convergence between ChatGPT and linguists in Exp3, using criteria from Sprouse et al. (2013). Note: Significant p-values are $p < .05$ and marginal p-values are $p \leq .1$. Significant Bayes factors are $BF > 3$, and marginal ones are $BF > 1$, in each direction.

The convergence rates inferred from different statistical methods were largely consistent. The mixed-effects logistic regression model indicated a convergence rate of 89% (131/148), while other methods suggested a rate of 88% (130/148). For comparison, the convergence rate between laypeople and linguists ranged from 91% (135/148) to 94% (139/148), contingent upon the applied statistical tests.

|  | One-tailed | Two-tailed | LME | Bayes factor |
| --- | --- | --- | --- | --- |
| Significant in the opposite direction | - | 3 | 2 | 3 |
| Marginal in the opposite direction | - | 0 | 0 | 0 |
| Non-significant in the opposite direction | - | 1 | 2 | 1 |
| Non-significant in the predicted direction | 7 | 5 | 7 | 5 |
| Marginal in the predicted direction | 2 | 0 | 2 | 0 |
| Significant in the predicted direction | 139 | 139 | 135 | 139 |

Table 10: Results of various statistical tests examining the convergence between laypeople and linguists, based on a re-analysis of human data from the FC task in Sprouse et al. (2013).

### 4.3 Discussion

In this experiment, we employed the FC task to assess how adeptly ChatGPT differentiates between grammatical sentences and their less-grammatical counterparts. While the raw percentages for correctly selecting the grammatical sentence across all trials were comparable between ChatGPT and human participants, the mixed model revealed that ChatGPT had a higher logit of selecting the correct answer than the human participants when accounting for the unique characteristics of experimental items. Figure 8 illustrates this: the diagonal line indicates identical probabilities for both human participants and ChatGPT in selecting the correct sentence. Data points above this diagonal indicate trials where ChatGPT was more accurate than human participants, and vice versa. Notably, more data points lie above the diagonal (457) than below it (168), signifying a higher accuracy rate for ChatGPT in many trials.

Interestingly, there was a noticeably weaker correlation between the judgments of ChatGPT and human participants in the FC task compared to the prior two tasks. This reduced correlation might be attributed to outliers where human participants vastly outperformed ChatGPT. Observing Figure 8, while a significant number of trials saw both ChatGPT and human participants achieving commendable accuracy (located in the upper right corner), there existed a cluster of trials where humans far outpaced ChatGPT. Such outliers clarify the observed phenomenon where the overall accuracy rate for humans across trials slightly surpassed that of ChatGPT, yet the trend inverted when random trial effects were accounted for. Table 11 provides examples of such outliers. In these particular trials, while humans predominantly favored the grammatical sentences (with accuracy rates >90%), ChatGPT, on the other hand, overwhelmingly opted for the ungrammatical ones (>90%).

| Trial ID | Grammatical Sentences | Ungrammatical Sentences |
| --- | --- | --- |
| 33.1.denDikken.71a.02 | Who the hell asked who out? | Who asked who the hell out? |
| 35.2.larson.44a.06 | A surfer cuter than my husband came along the beach. | A cuter surfer than my husband came along the beach. |
| 41.3.Landau.27b.05 | The teacher and principal spoke together after class. | The teacher spoke together after class. |
| 32.3.fanselow.58d.01 | There has been a man considered sick. | There has been considered a man sick. |
| 34.4.boskovic.3c.05 | They knew and we saw that Mark would skip work. | They knew and we saw Mark would skip work. |

Table 11: A selection of trials in Experiment 3 for which human participants had a much higher accuracy rate than ChatGPT (90% vs 10%).

These outliers point to specific grammatical constructs that ChatGPT struggles to grasp. However, it's challenging to identify any overarching pattern or generalization concerning these constructs. For instance, while the first sentence pair in Table 11 delves into the syntactical intricacies of the parenthetical element "the hell", the last pair revolves around the use of a complementizer in conjunct clauses. Each outlier seems to spotlight a distinct grammatical phenomenon, unrelated to the others.

Regarding the alignment between ChatGPT's responses and linguists' judgements, our findings indicate that different statistical methods rendered estimates approximating 89%. This infers that in 89% of the pairwise phenomena, ChatGPT's assessments concurred with linguists' perspectives. Conversely, in the remaining 11%, sentences deemed ungrammatical by linguists were interpreted as grammatical by ChatGPT and vice versa.

## 5. General Discussion
Expanding upon Sprouse and colleagues (2013), our study introduced ChatGPT as an AI counterpart for grammaticality judgment, seeking to determine the extent to which ChatGPT aligns with laypeople and linguists in their assessments of English sentence acceptability. In doing so, we contribute to the ongoing discourse surrounding the linguistic capabilities of cutting-edge large language models.

## 5.1 Alignment of Grammatical Knowledge Between ChatGPT and Laypeople

Regarding the alignment of grammatical knowledge between ChatGPT and laypeople, we found significant correlations among the two groups across all judgment tasks, though the strength of the correlation varied as a function of the task. Strong correlations were observed in the ME and LS tasks, while a weak correlation was observed in the FC task. This discrepancy, we believe, resulted from the task-specific idiosyncrasies.

In the ME and LS tasks, sentences of varying degrees of grammaticality were presented in a random order. The correlation between ChatGPT and human participants thus reflected their agreement in the "rankings" assigned to sentences based on their grammatical intuition. In the FC task, sentences were presented in pairs according to the membership of specific grammatical phenomena, and the responses were coded as "correct" or "incorrect" based on the pre-assigned grammaticality of the sentences. It is important to note that the concept of grammatical phenomenon and the categorical distinction of grammatical and ungrammatical sentences are predefined by linguists. Thus, the correlation between ChatGPT and human participants in the FC task reflects not so much the alignment in their grammaticality judgment of a spectrum of structures, rather it reflects their alignment in the degree in which their judgments fit in pre-defined categories.

Statistically speaking, for the ME and LS task, a high correlation between two groups implies that sentences receiving higher ratings in one group also receive higher ratings in the other group and vice versa. However, a general agreement in sentence ranking does not guarantee a high correlation in the FC task. This occurs because the FC task does not assess the grammaticality ranking across a range of sentences, but rather focuses on the relative grammaticality within sentence pairs. Even if both sentences in a pair were deemed highly acceptable, participants in the FC task would still have to make a choice between the two. The discrepancy between ChatGPT and human participants in this choosing process led to the decreased correlation in the FC task as compared to the ME and LS task. Table 11 in session 4.3 provides some examples to illustrate this point.

1a They knew and we saw that Mark would skip work.
1b They knew and we saw Mark would skip work.

It is likely that most readers would find the pair of sentences, 1a and 1b, generally more acceptable than the pair 2a and 2b:

2a Who the hell asked who out?
2b Who asked who the hell out?

The perceived naturalness of sentences 1a and 1b (or the perceived awkwardness in 2a and 2b) could stem from structural properties or their frequency of use. However, our discussion primarily centers on the factors contributing to the correlation estimates across tasks, rather than on what influences the grammaticality of the stimuli. In the rating tasks (ME and LS), a high correlation between ChatGPT and laypeople is observed if both groups generally agree that sentences like 1a and 1b sound better than 2a and 2b. However, a high correlation in the FC task requires that the two groups show the same preference between 1a and 1b, or between 2a and 2b; and in this study we did not observe a high correlation between ChatGPT and human participants in this aspect. While laypeople consistently preferred the first sentence over the second in both pairs, ChatGPT tended to show the opposite preference.

When shifting the attention from correlation to equivalence, we found that though ChatGPT and laypeople correlated in the ME task, their response patterns were noticeably distinct. Ratings from ChatGPT were more conservative than those from human participants in that grammatical sentences were rated lower by ChatGPT than by humans, while ungrammatical sentences were rated in the reversed trend. Aside from the ME task, differences between human participants and ChatGPT was minimal: In the LS task, the effect of participant type on rating scores was not statistically significant; as for the FC task, the difference between ChatGPT and humans, although existed, was almost negligible (89% vs 88%).

We believe that the wider range of human rating scores in the ME task results from variations in the number of response options and differences in language processing styles between humans and ChatGPT. Compared with the LS task and FC task where the response options form a limited set, the rating score that a participant could assign to a stimulus in the ME task is potentially unlimited (as long as it is proportional to the reference sentence). This feature of the ME task renders it more susceptible to the effect of individual differences in response style. Langsford et al. (2018) surveyed a series of tasks commonly used for grammaticality judgements, including ME, LS, and FC. They found that the between-participant reliability of the ME task being notably less than its within-participant reliability. Additionally, differences in language processing styles between humans and ChatGPT contribute to the narrower range of ChatGPT's ratings compared to those of humans in the ME task. Cai et al. (2023) demonstrated that ChatGPT tends to gloss over errors and normalize semantic and syntactic anomaly based on its knowledge of language use, which is derived from a large amount of training data. For example, when processing sentences like "The mother gave the candle the daughter", ChatGPT is more likely than humans to parse it as "The mother gave the candle to the daughter", because in its training data, human characters like "the daughter" is much more likely to be the recipient of an object like a "candle" than vice versa. It is possible that in the ME task, some ungrammatical sentences had been substantially normalized by ChatGPT, resulting in a higher rating of those sentences compared with human participants. This tendency might be more pronounced in the ME task, which offers a wider range of response options compared to the LS and FC tasks.

### 5.2 Alignment of Grammatical Knowledge Between ChatGPT and Linguists
In terms of the alignment in grammatical intuition between ChatGPT and linguists, the convergence estimates vary depending on the grammaticality judgement tasks and the analytical methods employed. As can be seen from Table 12, the highest convergence estimates were reported based on directionality which concerns only the difference in the mean between conditions (grammatical vs ungrammatical categories) but not the significance of the difference. On the other hand, linear mixed-effects model provided most conservative estimates for both the ME and LS tasks, whereas for the FC task, the estimates from the mixed-effects model were comparable to those from other statistical tests. This results in the estimated convergence rate ranging from 95% to 73% between ChatGPT and linguists.

Sprouse et al. (2013) adopted the estimate from the linear mixed effect model in the FC task as the point estimate of convergence. We agreed with this decision for two reasons. First of all, research on design sensitivity and statistical power of grammaticality judgement tasks has shown that among various experimental paradigms, FC task is the most sensitive task for detecting differences between two conditions and has the highest within-and-between participant reliability (Sprouse & Almeida, 2017; Langsford et al., 2018). Second, linear mixed effects model treats items as random effects, reducing false positive rate (Jaeger,

2008). By adopting the estimate from the linear mixed effect model in the FC task, the convergence rate in grammatical intuition between ChatGPT and linguists is 89%. Given that the distribution of the convergence rate is unknown, determining whether a convergence rate of 89% is notably high or low remains challenging. Nevertheless, considering that our reanalysis of Sprouse et al. (2013)[‡]'s human data yielded a 91% convergence rate between laypeople and linguists, we interpret the 89% ChatGPT-linguist rate as comparatively high.

|    | Directionality | One-tailed | Two-tailed | LME   | Bayes factor |
|----|----------------|------------|------------|-------|--------------|
| ME | 94             | 91/93      | 91         | 73/77 | 89/91        |
| LS | 95             | 95         | 94/95      | 75/80 | 94           |
| FC | 90             | 88         | 88         | 89    | 88           |

Table 12: A summary of convergence rates (in percentage) between ChatGPT and linguists across three grammaticality judgement tasks estimated by different analytical methods. In cells with slashes (/) the percentage on the left assumes that marginal results are non-significant; the percentage on the right assumes that marginal results are significant.

The observed correlations and distinctions across various judgment tasks shed light on the complexity of ChatGPT's linguistic competence. These findings have significant implications for natural language understanding and generation tasks, with ChatGPT demonstrating the potential to contribute to language-related applications and linguistic research. Further research is needed to explore the generalizability of ChatGPT's linguistic knowledge to other languages and investigate the impact of model size and fine-tuning on its grammatical judgments. Moreover, it is important to further investigate the specific linguistic constructs that pose challenges to ChatGPT's judgments. By addressing these areas, we can continue to refine and enhance the linguistic capabilities of large language models like ChatGPT, paving the way for more effective and nuanced natural language processing systems.

**6 Conclusion**
In conclusion, this research has undertaken a comprehensive investigation into the alignment of grammatical knowledge between ChatGPT, laypeople, and linguists, shedding light on the capabilities and limitations of AI-driven language models in approximating human linguistic intuitions. The findings indicate significant correlations between ChatGPT and both laypeople and linguists in various grammaticality judgment tasks. This study also reveals nuanced differences in response patterns, influenced significantly by the specific task paradigms employed. This study contributes to the ongoing discourse surrounding the linguistic capabilities of artificial intelligence and the nature of linguistic cognition in humans, calling for further exploration of the evolving landscape of linguistic cognition in humans and artificial intelligence.

---

[‡] We reanalyzed the human data from Sprouse et al. (2013) following their original procedure using most up-to-date statistical packages. Scripts are available from the OSF repository of this project: https://osf.io/crftu/

# Reference

Bates, D., Mächler, M., Bolker, B., Walker, S., Christensen, R. H., Singmann, H., Dai, B., Scheipl,F., Grothendieck, G., Green, P., Fox, J., Bauer, A., & Krivitsky, P. N. (2020). Package 'lme4'. R Package. Version 1.1-26. https://cran.rproject.org/web/packages/lme4/lme4.pdf

Bever, T. G. (1968). Associations to Stimulus-Response Theories of Language. In Dixon, T., and Horton, D. (eds.), Verbal Behavior and General Behavior Theory. 478-494.

Binz, M., & Schulz, E. (2023). Using cognitive psychology to understand GPT 3. Proceedings of the National Academy of Sciences, 120(6), e2218523120.

Brown, T., Mann, B., Ryder, N., Subbiah, M., Kaplan, J. D., Dhariwal, P., ... & Amodei, D. (2020). Language models are few-shot learners. *Advances in neural information processing systems*, 33, 1877-1901.

Brunet-Gouet, E., Vidal, N., & Roux, P. (2023). Do conversational agents have a theory of mind? A single case study of ChatGPT with the Hinting, False Beliefs and False Photographs, and Strange Stories paradigms. *HAL Open Science*, https://hal.science/hal-03991530/

Bürkner, P. C. (2017). brms: An R package for Bayesian multilevel models using Stan. Journal of statistical software, 80, 1-28.

Cai, Z. G., Haslett, D. A., Duan, X., Wang, S., & Pickering, M. J. (2023). Does ChatGPT resemble humans in language use? *arXiv preprint* arXiv:2303.08014.

Chomsky, N. (1980). Rules and representations. Behavioral and brain sciences, 3(1), 1-15.

Chomsky, N. (1986): Knowledge of Language: Its Nature, Origin, and Use, New York: Praeger Publishers.

Chomsky, N., (1965). Aspects of the Theory of Syntax. MIT Press, Cambridge, MA.

Chomsky, N.Noam, Ian Roberts I., & Jeffrey Watumull, .J. (2023). Noam Chomsky: The False Promise of ChatGPT. *The New York Times*. https://www.nytimes.com/2023/03/08/opinion/noam-chomsky-chatgpt-ai.html.

Cohen, J. (1988). Statistical power analysis for the behavioral sciences (2nd ed.). Hillside, NJ: Lawrence Erlbaum Associates.

Cohen, J. (1992). A power primer. Psychological Bulletin, 112, 155–159. doi:10.1037/0033-2909.112.1.155

Davies, W. D., & Kaplan, T. I. (1998). Native speaker vs. L2 learner grammaticality judgements. Applied linguistics, 19(2), 183-203.

Devitt, M. (2006). Intuitions in linguistics. British Journal for the Philosophy of Science, 57, 481–513

Ferreira, F. (2005). Psycholinguistics, formal grammars, and cognitive science. The Linguistic Review 22. 365–80.

Fodor, J. (1981). Introduction: Some notes on what linguistics is about. In The Language and Thought Series. Harvard University Press. 197-207.

Futrell, R., Wilcox, E., Morita, T., Qian, P., Ballesteros, M., & Levy, R. (2019). Neural language models as psycholinguistic subjects: Representations of syntactic state. *arXiv preprint* arXiv:1903.03260.

Graves, C., Katz, J. J., Nishiyama, Y., Soames, S., Stecker, R., & Tovey, P. (1973). Tacit knowledge. The Journal of Philosophy, 70(11), 318-330.

Hill, A. A. (1961). Grammaticality. Word, 17(1), 1-10.

Jaeger, T.F., 2008. Categorical data analysis: away from ANOVAs (transformation or not) and towards logit mixed models. Journal of Memory and Language 59, 434--446.

Jiao, W., Wang, W., Huang, J. T., Wang, X., & Tu, Z. (2023). Is ChatGPT a good translator? A preliminary study. *arXiv preprint* arXiv preprint arXiv:2301.08745.

Kosinski, M. (2023). Theory of mind may have spontaneously emerged in large language models. *arXiv preprint* arXiv preprint arXiv:2302.02083.

Langsford, S., Perfors, A., Hendrickson, A. T., Kennedy, L. A., & Navarro, D. J. (2018). Quantifying sentence acceptability measures: Reliability, bias, and variability. . Glossa: a journal of general linguistics 3(1): 37. 1–34, DOI: https://doi.org/10.5334/gjgl.396

Lyons, J. (1968). Introduction to theoretical linguistics. Harvard University Press, Cambridge, Massachusetts.

Mahowald, K., Fedorenko, E., Piantadosi, S. T., & Gibson, E. (2013). Info/information theory: Speakers choose shorter words in predictive contexts. *Cognition*, 126(2), 313-318.

Mandell, P. B. (1999). On the reliability of grammaticality judgement tests in second language acquisition research. Second Language Research, 15(1), 73-99.

Morey, R. D., Rouder, J. N., Jamil, T., Urbanek, S., Forner, K., & Ly, A. (2022). Package 'BayesFactor'. https://www.icesi.co/CRAN/web/packages/BayesFactor/BayesFactor.pdf

Ortega-Martín, M., García-Sierra, Ó., Ardoiz, A., Álvarez, J., Armenteros, J. C., & Alonso, A. (2023). Linguistic ambiguity analysis in ChatGPT. *arXiv preprint* arXiv:2302.06426.

Piantadosi, S. T. (2023). Modern language models refute Chomsky's approach to language. *Lingbuzz Preprint,* lingbuzz/007180.Piantadosi, S. T. (2023). Modern language models refute Chomsky's approach to language. file:///Users/qiuzhuang/Downloads/piantadosi_23_Modern-lang.2.pdf

Riemer, N. (2009). Grammaticality as evidence and as prediction in a Galilean linguistics. Language Sciences, 31(5), 612-633.

Rouder, J.N., Speckman, P.L., Sun, D., Morey, R.D., and Iverson, G. (2009). Bayesian t-tests for accepting and rejecting the null hypothesis. Psychon Bull Rev, 16:225–237.

Saussure, F. (1916). Cours de linguistique générale Paris: Payot.

Schütze, C. T. (1996). The empirical base of linguistics. Chicago: University of Chicago Press.

Sprouse, J. & Almeida, D., (2017) "Design sensitivity and statistical power in acceptability judgment experiments", Glossa: a journal of general linguistics 2(1): 14. doi: https://doi.org/10.5334/gjgl.236

Sprouse, J., Schütze, C. T., & Almeida, D. (2013). A comparison of informal and formal acceptability judgments using a random sample from Linguistic Inquiry 2001–2010. Lingua, 134, 219-248.

Sprouse, J., Wagers, M. W., & Phillips, C. (2013). Deriving competing predictions from grammatical approaches and reductionist approaches to island effects. In Jon Sprouse & Nobert Hornstein (Eds.) Experimental Syntax and Island Effects. Cambridge: Cambridge University Press. 21-41.

Van der Lely, H. K., Jones, M., & Marshall, C. R. (2011). Who did Buzz see someone? Grammaticality judgement of wh-questions in typically developing children and children with Grammatical-SLI. Lingua, 121(3), 408-422.

Van Schijndel, M., & Linzen, T. (2018). Modeling garden path effects without explicit hierarchical syntax. In *Proceedings of the 40th Annual Meeting of the Cognitive Science Society*.

Wilcox, E. G., Futrell, R., & Levy, R. (2022). Using computational models to test syntactic learnability. *Linguistic Inquiry*, 1-88.

Wu, H., Wang, W., Wan, Y., Jiao, W., & Lyu, M. (2023). Chatgpt or grammarly? evaluating chatgpt on grammatical error correction benchmark. arXiv preprint arXiv:2303.13648.